\documentclass[conference]{IEEEtran}

\usepackage{cite}
\usepackage{url}
\usepackage[pdftex]{graphicx}

\usepackage{tikz}
\usepackage{pgfplots}
\pgfplotsset{compat=newest}
\usetikzlibrary{pgfplots.groupplots}

\usepackage[cmex10]{amsmath}
\usepackage{amsfonts}
\usepackage{amsthm}

\newtheorem{requirement}{Requirement}

\usepackage{algorithm}
\usepackage{algorithmicx}
\usepackage{algpseudocode}
\usepackage[algo2e,ruled,vlined,linesnumbered]{algorithm2e}
\usepackage{xspace}
\usepackage{soul}

\usepackage{dblfloatfix}

\usepackage{multirow}

\graphicspath{{pic/}}

\allowdisplaybreaks

\IEEEoverridecommandlockouts

\newcommand{\ignore}[1]{}

\newcommand{\N}{\mathbb{N}}
\newcommand{\R}{\mathbb{R}}

\renewcommand{\epsilon}{\varepsilon}

\renewcommand{\Pr}{\mathbb{P}}

\DeclareMathOperator*{\argmin}{arg\,min}
\DeclareMathOperator*{\argmax}{arg\,max}

\newcommand{\pmn}{p_\text{min}}

\newcommand{\onemax}{\textsc{OneMax}\xspace}

\newcommand{\OneMax}{\textsc{OneMax}\xspace}
\newcommand{\OM}{\textsc{Om}\xspace}

\newcommand{\leadingones}{\textsc{LeadingOnes}\xspace}

\newcommand{\Jump}{\textsc{Jump}\xspace}

\newcommand{\Ruggedness}{\textsc{Ruggedness}\xspace}

\newcommand{\shmut}{shift mutation\xspace}

\newcommand{\oplab}{$(1 + \lambda) \text{ EA}(A,b)$\xspace}

\newcommand{\opl}{$(1 + \lambda)$~EA\xspace}

\newcommand{\hqea}{$(1 + \lambda)$~HQEA\xspace}

\newcommand{\oplplot}{$(1+\lambda)$}

\newcommand{\hybridplot}{HQEA\xspace}
\newcommand{\hybrid}{\hqea}

\newcommand{\ABplot}{$(A, b)$\xspace}
\newcommand{\AB}{\oplab}

\newcommand{\tworate}{$(1+\lambda) \text{ EA}_{r/2,2r}$\xspace}
\newcommand{\rateplot}{2-rate\xspace}
\newcommand{\rate}{\tworate}

\pgfplotsset{
    every non boxed y axis/.style={}
}

\pgfplotscreateplotcyclelist{colorcycle}{%
red!80!black,mark=*,mark size=1.3pt\\
blue,mark=square*,mark size=1.3pt\\
cyan!80!black,mark=*,mark size=1.3pt\\
violet!80!black,mark=square*,mark size=1.3pt\\
green,mark=*,mark size=1.3pt\\
orange,mark=mark=square*,mark size=1.3pt\\
yellow, mark=*,mark size=1.3pt\\
}

\pgfplotscreateplotcyclelist{noqcolorcycle}{%
red!80!black,mark=*,mark size=1.3pt\\
blue,mark=square*,mark size=1.3pt\\
cyan!80!black,mark=*,mark size=1.3pt\\
orange,mark=diamond*\\
}

\pgfplotscreateplotcyclelist{othercolorcycleinv}{%
cyan,mark=diamond*,mark size=1.3pt\\
blue,mark=diamond*,mark size=1.3pt\\
}

\pgfplotscreateplotcyclelist{othercolorcycle}{%
blue,mark=diamond*,mark size=1.3pt\\
cyan,mark=diamond*,mark size=1.3pt\\
blue,mark=diamond*,mark size=1.3pt\\
cyan,mark=diamond*,mark size=1.3pt\\
blue,mark=diamond*,mark size=1.3pt\\
cyan,mark=diamond*,mark size=1.3pt\\
blue,mark=diamond*,mark size=1.3pt\\
cyan,mark=diamond*,mark size=1.3pt\\
blue,mark=diamond*,mark size=1.3pt\\
cyan,mark=diamond*,mark size=1.3pt\\
blue,mark=diamond*,mark size=1.3pt\\
cyan,mark=diamond*,mark size=1.3pt\\
blue,mark=diamond*,mark size=1.3pt\\
cyan,mark=diamond*,mark size=1.3pt\\
blue,mark=diamond*,mark size=1.3pt\\
cyan,mark=diamond*,mark size=1.3pt\\
blue,mark=diamond*,mark size=1.3pt\\
cyan,mark=diamond*,mark size=1.3pt\\
blue,mark=diamond*,mark size=1.3pt\\
cyan,mark=diamond*,mark size=1.3pt\\
}

\newcommand{\plottwodiff}[4]{
\begin{tikzpicture}

\begin{axis}[name=ax1, title={$p_{\min} = 1/n^2$}, title style={at={(0.45,-0.3)}, anchor=north}, legend columns = 5, legend style={at={(1.03,1.03)}, anchor=south}, height 
= 0.218\textheight, width = 0.25\textwidth, enlarge x limits=false, enlarge y limits=false, xmode=log, ymode=log, xlabel = {$\lambda$}, log base x=2, grid = both, cycle 
list name=colorcycle, #3]
#1{#2}{sq}{strict}
\end{axis}

\begin{axis}[name=ax2, title={$p_{\min}=1/n$}, title style={at={(0.45,-0.3)}, anchor=north}, at={(ax1.south east)}, xshift=1cm, height = 0.218\textheight, width = 0.25\textwidth, 
enlarge x limits=false, enlarge y limits=false, xmode=log, ymode=log, xlabel = {$\lambda$}, log base x=2, grid = both,  cycle list name=colorcycle, #3, ylabel={}]
#1{#2}{}{strict}
\legend{};
\end{axis}

\end{tikzpicture}
}

\newcommand{\plottwo}[3]{
\plottwodiff{#1}{#2}{#3}{#1}
}

\newcommand{\addquartdevwithoptimum}[3]{
	
	\addplot plot[error bars/.cd, y dir=both, y explicit] table [x=lambda, y=average, y error=dev, col sep=comma] {data/#1/simple.csv};
	\addlegendentry {\footnotesize\oplplot};
	
	\addplot plot[error bars/.cd, y dir=both, y explicit] table [x=lambda, y=average, y error=dev, col sep=comma] {data/#1/tworate#2.csv};
	\addlegendentry {\footnotesize\rateplot};
	
	\addplot plot[error bars/.cd, y dir=both, y explicit] table [x=lambda, y=average, y error=dev, col sep=comma] {data/#1/AB#3#2.csv};
	\addlegendentry {\footnotesize\ABplot};
	
	\addplot plot[error bars/.cd, y dir=both, y explicit] table [x=lambda, y=average, y error=dev, col sep=comma] {data/#1/HQEA#3#2.csv};
	\addlegendentry {\footnotesize\hybridplot};
	
	\addplot plot table [x=lambda, y=time, col sep=comma] {data/#1/optimal.csv};
	\addlegendentry {\footnotesize Lower};
}

\newcommand{\LowerBoundPlot}[1]{\begin{figure*}[#1]
    \centering
    \begin{tabular}{cc}
    \plottwo{\addquartdevwithoptimum}{OneMax_100_100}{ylabel={Iterations}, ymax=2000, ytick = {100, 1000, 10000, 100000, 1000000, 10000000, 100000000}, yticklabels = {$10^2$, $10^3$, $10^4$, $10^5$, $10^6$, $10^7$, $10^8$}, xtick={2, 8, 32, 128, 512}, xticklabels={$2^1$, $2^3$, $2^5$, $2^7$, $2^9$}}
    &
    \plottwo{\addquartdevwithoptimum}{Ruggedness2_100_100}{ylabel={Iterations}, ymax=200000000, ytick = {100, 1000, 10000, 100000, 1000000, 10000000, 100000000}, yticklabels = {$10^2$, $10^3$, $10^4$, $10^5$, $10^6$, $10^7$, $10^8$}, xtick={2, 8, 32, 128, 512}, xticklabels={$2^1$, $2^3$, $2^5$, $2^7$, $2^9$}}
    \\
    \OneMax & \Ruggedness
    \end{tabular}
    \caption{Lower bounds for the expected running times of the \opl using binomial distribution with varying $p$ for mutation,
             compared with the performance of various parameter control methods averaged over 100 independent runs. Problem size is $n=100$ for all four plots.
             Legend entries are: \oplplot: the \opl with fixed $p=1/n$, \rateplot: the \tworate, \ABplot: the \oplab, \hybridplot: the \hybrid.}
    \label{fig:lower-bounds}
\end{figure*}}

\newcommand{\BigHeatmapPlot}{
    \centering
    \begin{tabular}{cc}
    \begin{tikzpicture}
        \begin{axis}[enlargelimits=false, axis on top, width=0.45\linewidth, height=0.28\textheight,
                     xlabel={Fitness}, ylabel={Scaled mutation rate: $p \cdot n$},
                     ymode=log, point meta min=0, point meta max=1] 
            \addplot graphics [xmin=0, xmax=99, ymin=0.00919, ymax=100, includegraphics={trim=0 16 0 0,clip}]{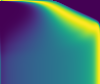};
        \end{axis}
    \end{tikzpicture}
    &
        \begin{tikzpicture}
    	\begin{axis}[enlargelimits=false, axis on top, width=0.45\linewidth, height=0.28\textheight,
    		xlabel={Fitness}, ylabel={Scaled mutation rate: $p \cdot n$},
    		ymode=log, point meta min=0, point meta max=1, colorbar, colormap name=viridis, colorbar/width=2.5mm]
    		\addplot graphics [xmin=0, xmax=99, ymin=0.01, ymax=100]{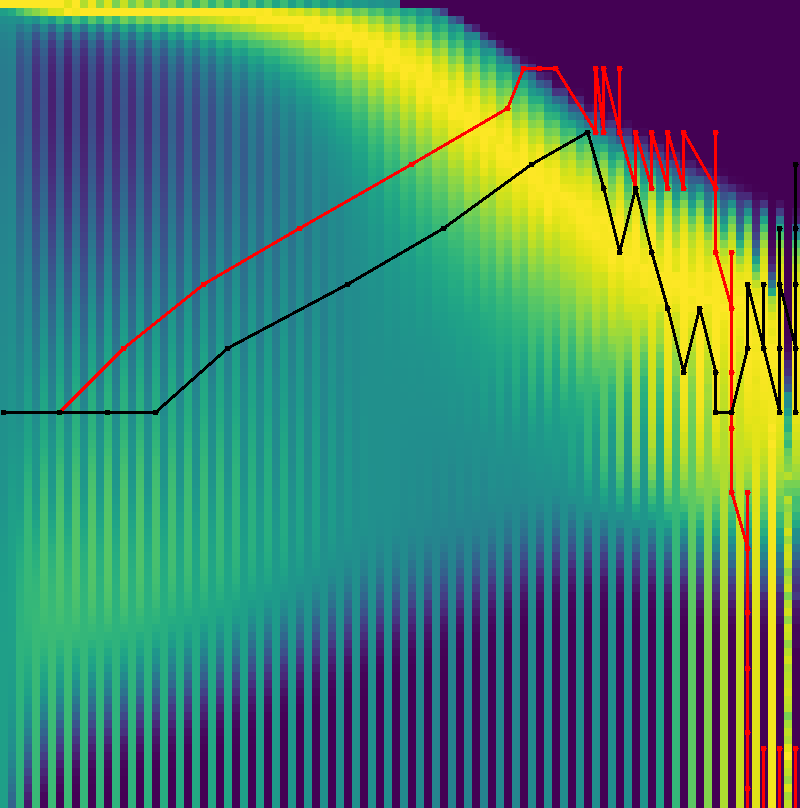};
    	\end{axis}
    \end{tikzpicture}\\
    \phantom{OnenMM}\OneMax & \phantom{n}\Ruggedness
    \end{tabular}
}

\begin{document}

\title{Blending Dynamic Programming with Monte Carlo Simulation for Bounding the Running Time of Evolutionary Algorithms}
\author{
\IEEEauthorblockN{Kirill Antonov}
\IEEEauthorblockA{
ITMO University,\\
Saint Petersburg, Russia
}
\and
\IEEEauthorblockN{Maxim Buzdalov}
\IEEEauthorblockA{
ITMO University,\\
Saint Petersburg, Russia
}
\and
\IEEEauthorblockN{Arina Buzdalova}
\IEEEauthorblockA{
ITMO University,\\
Saint Petersburg, Russia
}
\and
\IEEEauthorblockN{Carola Doerr}
\IEEEauthorblockA{
Sorbonne Universit{\'e},\\
CNRS, LIP6, Paris, France
}
}

\maketitle

\begin{abstract}
With the goal to provide absolute lower bounds for the best possible running times that can be achieved by $(1+\lambda)$-type search heuristics on common benchmark problems, we recently suggested a dynamic programming approach that computes optimal expected running times and the regret values inferred when deviating from the optimal parameter choice. 

Our previous work is restricted to problems for which transition probabilities between different states can be expressed by relatively simple mathematical expressions. With the goal to cover broader sets of problems, we suggest in this work an extension of the dynamic programming approach to settings in which the transition probabilities cannot necessarily be computed exactly, but in which they can be approximated numerically, up to arbitrary precision, by Monte Carlo sampling. 

We apply our hybrid Monte Carlo dynamic programming approach to a concatenated jump function and demonstrate how the obtained bounds can be used to gain a deeper understanding into parameter control schemes. 
\end{abstract}

\section{Introduction}

Running time analysis of evolutionary algorithms and other search heuristics contributes to our understanding of black-box optimization not only by providing insights into the basic working principles that drive algorithms' performance, but also by providing \emph{lower bounds} for the performance of broad classes of black-box approaches~\cite{DoerrN20,Doerr20chapter}. Typically expressed as black-box complexity bounds, these lower bounds can be seen as a baseline against which we can compare state-of-the-art solvers, with the goal to quantify the potential of further algorithm development. By comparing black-box complexity bounds for different classes of algorithms, we obtain insight into the impact of certain algorithm features, such as their degree of parallelism~\cite{ParallelBBC14,LehreS20parallelBBC}, their memory~\cite{DrosteJW06,DoerrW14memory}, their selection principles~\cite{DoerrL17ECJ}, properties of their sampling strategies~\cite{LehreW12,ABB,DoerrKLW13}, etc. More refined lower bounds can be obtained by specifying a family of algorithms, the available choices for configuring it, and by studying the dependency of the running time on the choices of these configurable components. This has been classically studied in the asymptotic sense~\cite{Witt13j,Sudholt13,GiessenW17,OlivetoSW20}, however, some recent works provide tight bounds also for concrete problem dimensions~\cite{ChicanoSWA15precise,GiessenW18}.

Most of these works assume a \emph{static} setting of the parameters. In practice, however, it is well know that \emph{dynamic parameter settings} can greatly improve the efficiency of evolutionary algorithms~\cite{KarafotiasHE15,AletiM16,EibenHM99}. Precise lower bounds for algorithms with dynamic parameter settings are rare, see~\cite{DoerrD18chapter} for a survey of rigorous results for algorithms with dynamic parameter choices and~\cite{BottcherDN10,DoerrDY20,LissovoiOW20} for a few exceptions that provide precise lower bounds for algorithms with dynamic parameter settings. 

Proving precise lower bounds for algorithms with dynamic parameter choices is challenging for several reasons: 
(i) lower bounds require to make statements about \emph{all} possible settings (as opposed to upper bounds, in which one concrete parameter setting is analyzed), 
(ii) the (often very complex and trajectory-based) dependency of the parameters on the state of the algorithm needs to be taken into account, and 
(iii) all bounds need to be very tight to obtain an overall bound that is precise enough to give meaningful results for concrete problem dimensions (and not just for the asymptotic behavior). 
In lack of precise methods to derive optimal algorithm behavior, several author teams have resorted to compute actions that maximize the step-wise progress, in hope that this ``drift-maximization'' is close to the true optimal parameter setting policy. This step-wise progress maximization was either done \emph{exactly}~\cite{Back92} or via \emph{Monte Carlo simulations}~\cite{FialhoCSS08}. It is known, however, that even for the \onemax problem with perfect fitness-distance correlation, the drift-maximization strategy is only close to being optimal~\cite{DoerrDY20}, but not strictly optimal~\cite{BuskulicD21,BuzdalovD20}. A different approach to compute precise lower bounds for algorithms with dynamic parameter settings was therefore taken in~\cite{BuskulicD21}. For the problem of minimizing the expected running time of (1+1)-type evolutionary algorithms on \onemax, the authors first derive an exact method to compute the optimal dynamic parameter choices and then evaluate this formula for concrete problem dimensions using a dynamic programming approach. 
The work was greatly extended in~\cite{BuzdalovD20}, by considering different $(1+\lambda)$-type heuristics and by not only computing the optimal parameter choices at each state, but also the cost of deviating from it. 

\textbf{Our contribution:} We demonstrate in this work how the approach taken in~\cite{BuskulicD21,BuzdalovD20} can be extended to settings in which we do not necessarily have a closed-form expression from which we can derive exact parameter settings. Our overall approach is still based on dynamic programming, but we replace the exact formulae that model the probabilities to transition between different states by Monte Carlo simulations. That is, we take a similar approach as in~\cite{FialhoCSS08}, but we do not simply maximize the expected drift, but use these approximated probabilities to derive strategies that minimize the expected running time -- the quantity that most running time analyses focus on. 

We apply our approach to the \Ruggedness function suggested in~\cite{DoerrYHWSB20}. \Ruggedness extends the classical jump function~\cite{DrosteJW02} by introducing several small jumps that the algorithms need to make in order to find the optimal solution, see Section~\ref{sec:ruggedness} for a formal definition. 
We have chosen this example because 
in~\cite{BuzdalovaDR20} we observed that classical parameter control schemes essentially fail on \Ruggedness, in the sense that they show worse performance than the $(1+\lambda)$ evolutionary algorithm (EA) with static mutation rate $1/n$ (with $n$ denoting the dimension of the problem). Interestingly, a better performance is obtained when mutation rates are capped at $1/n$ instead of $1/n^2$; see Figure~\ref{fig:lower-bounds} for an illustration. That is, despite having less flexibility, the algorithms perform better. High-performing parameter control mechanisms should not suffer from more flexibility, and we therefore see this example as an interesting use-case to identify deficiencies of state-of-the-art parameter control techniques, and to develop appropriate remedies to mitigate these. 

We suggested in~\cite{BuzdalovD20} heatmaps to illustrate the behavior of parameter control schemes. While indeed being very useful to bound the regret (i.e., the cost of deviating from the optimal parameter setting) of a single iteration, these heatmaps turned out to be suboptimal when analyzing the whole search trajectory. We therefore suggest in this work to complement them with plots that illustrate the regret per iteration. With these plots, one can now make a clear distinction between quick cheap ``hiccups'' and long-term harmful deviations.

\textbf{Implications for Ruggedness:} The application of our method to \Ruggedness shows that the \opl with static mutation rate $1/n$ is around 50-75\% worse than the \opl variant using optimal dynamic mutation rates. We also illustrate the usefulness of our regret plots by analyzing why the \AB with generalized 1/5-th success rule, shown efficient for other problems in~\cite{DoerrW18}, 
 fails on this problem. 

\textbf{Impact and limitations of our work:} While the precise values of the lower bound and of the individual regrets for \Ruggedness may be of interest to a rather specialized community only, we believe that our general approach of hybridizing Monte Carlo simulation with dynamic programming should be of interest to everyone willing to derive precise lower bounds against which sampling-based optimization algorithms can be compared. 

At present, our work is restricted to settings in which states are never visited twice. In practice, this limits our work to greedy (``elitist'') algorithms. Extensions to algorithms and/or problems in which states may be visited more than once forms one of the most challenging directions for future work, on which we comment in more detail in Section~\ref{sec:generalization}.

\textbf{Structure of the paper:} 
The \opl framework and its main variants studied in~\cite{BuzdalovaDR20} are introduced in Section~\ref{sec:opl}. In Section~\ref{sec:algo} we describe our hybrid dynamic programming approach using Monte Carlo simulation. In Section~\ref{sec:usecase} we present the results of this approach applied to derive the dynamic parameter settings which minimize the expected running time of the \opl on \Ruggedness. We comment on further extensions of our approach in Section~\ref{sec:generalization}. Section~\ref{sec:conclusions} concludes the paper.



\section{$(1+\lambda)$ Evolutionary Algorithms}
\label{sec:opl}


\begin{algorithm2e}[!t]
	\caption{A family of $(1+\lambda)$-type algorithms}
	\label{alg:opl}
	\KwData{$n$: problem size; $f: \{0,1\}^n \to \R$: function to maximize; $\lambda$: population size; $\mathcal{D}(p)$: a family of parameterized distributions over $[0..n]$}
	Sample parent $x \in \{0,1\}^n$ uniformly at random\;
	\For{$t \gets 1, 2, \ldots$}{
		\For{$i \in [1..\lambda]$}{
				Choose a distribution parameter $p^t_i$\;
		    Sample $k_i \sim \mathcal{D}(p^t_i)$, the number of bits to flip\;
		    Create $y_i$ by flipping $k_i$ bits in $x$ chosen uniformly at random without replacement\;
		}
		Select $x \gets \argmax_{z \in \{x, y_1, \ldots, y_{\lambda}\}} f(z)$\label{line:selection}\;
    }
\end{algorithm2e}

We present in Algorithm~\ref{alg:opl} a fairly general family of $(1+\lambda)$~EAs, for the context of maximizing a ``fitness'' function $f: \{0,1\}^n \to \R$ defined on bit strings. Algorithms covered by this framework are initialized by selecting their first solution candidate uniformly at random. In each iteration, $\lambda$ ``offspring'' are sampled, by modifying the best-so-far solution $x$. The modification is done by unary unbiased variation operators~\cite{LehreW12}, which --according to a characterization provided in~\cite{DoerrDY20} -- are defined by fixing a distribution $\mathcal{D}$ over the set of possible mutation strengths~$k \in [0..n]:=\{0,1,2,\ldots,n\}$. 
The operator first samples a mutation strength $k$ from this distribution and then flips $k$ uniformly selected, pairwise different bits in $x$.\footnote{Examples: The two most commonly used unary unbiased variation operators are 1-bit-flips and standard bit mutation (SBM). 1-bit-flips, as used within Randomized Local Search (RLS), correspond to the 1-point distribution $\Pr[k=1]=1$, whereas SBM (used by many evolutionary algorithms) is exactly the operator characterized by the binomial distribution $\mathcal{B}(n,p)$.} The best-so-far solution ``survives'' and determines the center of variation in the following iteration. Ties can be broken arbitrarily; in practice and in the remainder of this work we assume uniform selection among the best offspring (if at least one of them is at least as good as the parent). 

We do not specify in Algorithm~\ref{alg:opl} a stopping criterion, because we are -- in line with the majority of running time analysis papers -- interested in the \emph{expected optimization time,} i.e., the average number of function evaluation that an algorithm performs until it evaluates for the first time a solution~$x^*$ with $f(x^*) \ge f(y)$ for all $y \in \{0,1\}^n$. 


In~\cite{BuzdalovaDR20}, the following \opl variants were studied:

\textbf{(1) Shift \opl:} The standard \opl with mutation rate~$p$ is Algorithm~\ref{alg:opl} with $\mathcal{D}(p^t_i)=\mathcal{B}(n,p)$ for all $t \in \N$, $i \in [1..\lambda]$. It was argued in~\cite{practice-aware,pinto-doerr-crossover-ppsn18} that for $(1+\lambda)$ schemes this operator should not be used in practice, since it assigns probability mass to flipping 0 bits, which cannot advance the search. Two strategies were suggested to mitigate this unfavorable behavior: ``shifting'' all probability mass from $k=0$ to $k=1$ and ``resampling'' $k$ until $k>0$. We use the shift strategy in all algorithms considered in this work, i.e.,  
we use the mutation operator characterized by the distribution $\mathcal{B}_{0 \rightarrow 1}(n,p)$,
which has $\Pr[\ell=1\mid \ell\sim\mathcal{B}_{0 \rightarrow 1}(n,p)] = \Pr[\ell\le 1 \mid \ell \sim\mathcal{B}(n,p)]$ and
$\Pr[\ell=k\mid\ell\sim\mathcal{B}_{0 \rightarrow 1}(n,p)] = \Pr[\ell=k\mid\ell\sim\mathcal{B}(n,p)]$ for $k \ge 2$.
Where $p$ is not specified, we tacitly assume $p=1/n$. 

\textbf{(2) \AB:} This algorithm, analyzed in~\cite{DoerrW18}, uses a multiplicative parameter update scheme inspired by the 1/5-th success rule~\cite{Rechenberg}. Our version uses \shmut (as introduced above) instead of resampling mutation, but the update of $p$ remains the same as in~\cite{DoerrW18} with $A=2$ and $b=1/2$. That is, $p$ is initialized as $1/n$. At the end of each iteration $p$ is updated to $2p$ if $\max\{f(y_1), \ldots, f(y_{\lambda}) \} \ge f(x)$ and to $p/2$ otherwise. 

\textbf{(3) \rate:} This algorithm was suggested in~\cite{DoerrGWY19}. It creates  $y_1, \ldots, y_{\frac{\lambda}{2}}$ by the \shmut with mutation rate $p/2$ and the other offspring, $y_{\frac{\lambda}{2}+1}, \ldots, y_{\lambda}$ by the \shmut with mutation rate $2p$.
Let $f_{h} = \max\{f(y_1), \ldots, f(y_{\frac{\lambda}{2}})\}$ and $f_{d} = \max\{f(y_{\frac{\lambda}{2}+1}), \ldots, f(y_{\lambda})\}$.
If $f_{h} > f_{d}$, $p$ is updated to $p/2$ with probability $3/4$ and to $2p$ otherwise. 
If $f_{d} > f_{h}$, $p$ is updated to $2p$ with probability $3/4$ and to $p/2$ otherwise. 
If $f_{d} = f_{h}$, $p$ is updated to either $p/2$ or $2p$ equiprobably.

\textbf{(4) \hqea:} This algorithm was suggested in~\cite{BuzdalovaDR20}. It hybridizes reinforcement learning with the multiplicative update used within the \AB. We cannot present details, for reasons of space, and refer the interested reader to~\cite{BuzdalovaDR20} for details. For this work, it is interesting to note that the algorithm also uses \shmut, with a mutation rate $p$ that is updated after each iteration. All $\lambda$ offspring are sampled from the same distribution.  

\textbf{Capping of the mutation rate $p$:} 
An advantage of the \shmut operator is that it converges to the 1-bit-flip operator when $p \rightarrow 0$. It can therefore be used to interpolate between global and local search. However, we typically want to maintain \emph{some} probability of escaping local optima. In practice, $p$ is therefore often capped to remain within an interval $[p_{\min},p_{\max}]$. In our work, when considering parameter control methods, we fix $p_{\max}=1/2$. As discussed in the introduction, $p_{\min}$ can have a decisive influence on the efficiency of the algorithms, as can be seen in the example on the right side of Figure~\ref{fig:lower-bounds}. We study two values, $p_{\min}=1/n$ and $p_{\min}=1/n^2$.

\textbf{Scope of our work:} We will focus in the following on computing a lower bound for the family of algorithms that follow Algorithm~\ref{alg:opl}, but which use identical distributions to sample the $\lambda$ offspring (i.e., we require that $\mathcal{D}(p^t_i)=\mathcal{D}(p^t_j)$ for all $i,j \in [\lambda]$). We also restrict the algorithms to those in which $\mathcal{D}(p^t)$ (using the previous convention, we will from now on drop the index $i$) may depend on $n$ and on $f(x)$, but not on $x$,\footnote{This assumption is irrelevant for our use-case, but the structure of $x$ can have an important impact in general, as is easily seen for the \leadingones function, for which $f(0\ldots 0)=f(01\ldots 1)=0$, but the optimal mutation rate for $(0\ldots 0)$ is one, whereas the optimal one for $(01\ldots 1)$ is $p_{\min}$.} nor on $t$, nor on any other information.


\section{Algorithm description}
\label{sec:algo}

In this section we describe our algorithm that computes good choices of distribution parameters $p$
for each possible fitness value of the parent.

\subsection{High-Level Description}

Our algorithm is based on dynamic programming. We iterate over the possible fitness values, starting from the second best and continuing towards the smallest value.
For each fitness value we aim at finding the best possible parameter value that minimizes the remaining expected running time, as well as the remaining time itself,
assuming that the \opl will subsequently choose the optimal parameter values for all better fitness values.
As a side effect we also compute the remaining expected running times for a number of parameter values, which will be later useful to evaluate regrets associated with
particular parameter control schemes.

While doing it, we assume, similarly to~\cite{BuskulicD21,BuzdalovD20}, that for all higher fitness values the best possible expected running times are already computed.
However, since we aim at dealing with various fitness functions, we use the Monte Carlo approach to approximate transition probabilities instead.

\subsection{Requirements for the Fitness Function}

Before diving into details, we discuss the limitation of this algorithm first.
Our main limitation follows from our assumption that, apart from the problem size, the optimal choice of the distribution parameter $p$ depends
on the fitness value exclusively. Hence, if two individuals have the same fitness but different structure, it may result in different transition probabilities
to higher fitness values. Failing to account for that may result in both overestimation and underestimation of running times.
For this reason, we formulate our requirement to the fitness function as follows.

\begin{requirement}
For any two individuals $x_1$ and $x_2$, such that $f(x_1) = f(x_2)$:\label{algorithm-requirement}
\begin{itemize}
    \item either both $x_1$ and $x_2$ shall be optima;
    \item or for each fitness value $f' > f(x_1)$, assuming $Y = \{ y \mid f(y) = f' \}$ is a set of bit strings with the fitness $f'$,
          there shall exist a bijective mapping $\pi : Y \to Y$ such that
          such that for each $y \in Y$ the following transition probabilities shall be equal: $\Pr[x_1 \to y] = \Pr[x_2 \to \pi(y)]$.
\end{itemize}
\end{requirement}

When applied inductively, this requirement informally means that the fitness value unambiguously determines the further stochastic behavior of the evolutionary algorithm.

Note that some popular benchmark fitness functions, such as \OneMax, as well as the function \Ruggedness used in this study,
satisfy this requirement. Some other functions satisfy it only partially: for instance, with certain definitions of the \Jump function,
the individuals that form the small fitness valley may have the same fitness but be structurally different.
However, in this particular case the probability that the parent of the \opl becomes such an individual during the optimization run is overwhelmingly small,
so we can still apply our algorithm and get the results that are imprecise only up to a factor whose difference from one is negligible.

\subsection{Detailed Description}

\begin{algorithm2e}[!t]
    \caption{High-level outline of the hybrid algo.}
    \label{alg:optimalAlg}

    $f_{\min}, f_{\max} \gets$ minimum and maximum fitness values\;	
    Initialize optimal times: $T^{*}_{f_{\max}} \gets 0$\; 
    \For{$f \gets f_{\max}-1, \dots, f_{\min}$}{
        \For{$ p \in \{p^{(f)}_1, p^{(f)}_2, \dots, p^{(f)}_{m_f}\} $}{
            Compute approximate probabilities $(\tilde{p}_i)_{i=0,1,\ldots}$ of increasing fitness by $i$ with mutation rate $p$ using the Monte Carlo approach\;
            $T_{f,p} \gets \dfrac{1}{1-\tilde{p}_0} \left(1 + \sum_{i>f} T_{i} \cdot \tilde{p}_{(i-f)}\right)$\; \label{dpTransitions}
        }
        Store optimal time: $T^{*}_f \gets \min_{p}(T_{f,p})$\;
        Store optimal rate: $P^{\text{opt}}_f \gets \argmin_{p}(T_{f,p})$\;
    }
    \Return $\{P^{\text{opt}}, T^{*}, T\}$
\end{algorithm2e}

The high-level pseudocode of the proposed algorithm is given in Algorithm~\ref{alg:optimalAlg}.

We maintain $T^*_f$, our approximation of the optimal expected remaining time starting at fitness $f$ until the optimum is hit. This value for the maximum fitness, $f_{\max}$ that corresponds to the optimum, is obviously zero: $T^*_{f_{\max}} = 0$. We compute the remaining values starting from $f_{\max} - 1$ and stepping down until the minimum fitness value $f_{\min}$ is processed. Since the algorithms from the \opl family are elitist, $T^*_f$ depends only on $T^*_{f'}$ with $f' > f$, but not on other entries.

To evaluate a particular $T^*_f$, we evaluate the expected remaining times $T_{f,p}$ starting at fitness $f$ until the optimum is hit, provided that (i) while the parent's fitness is $f$, the \opl uses the mutation rate $p$, and (ii) when the parent's fitness is updated, the \opl uses the previously computed optimal mutation rates. Since we aim at an easy-to-compute approximation scheme that does not depend on the structure of the fitness function, we employ the following simplifications.
\begin{itemize}
    \item Instead of testing all possible $p$, which is computationally infeasible, we use a finite set of mutation rates $\{ p^{(f)}_1, \ldots, p^{(f)}_{m_f} \}$ that may depend on $f$. To obtain a good approximation, this set shall be dense enough around the assumed optimal mutation rate, but in the same time it should have a small enough size so that the computation finishes in affordable times. We detail our choices of these sets later in Section~\ref{sec:rate-sets}.
    \item Instead of analytically computing transition probabilities of the \opl from the current fitness $f$ to each larger fitness $f'$ (which may be very hard, error-prone and computationally demanding for certain problems) we use the Monte Carlo approach, which amounts to simulation of a part of the run of the \opl.
\end{itemize}

The Monte Carlo evaluation of $T_{f,p}$ is performed as follows.
\begin{enumerate}
    \item Choose a large number of modeled iterations, $N_I$.
    \item Choose a large number $N_T$ of \emph{successful} iterations (the iterations that resulted in a strict fitness increase) to wait for: if the transitions are easy enough, $N_T$ successes already make a good estimation of the majority of likely transitions, so we do not execute all $N_I$ iterations and hence save computational resources.
    \item Find an individual with fitness $f$ and set it as the parent individual for the next iterations of the \opl.
    \item Perform $N_I$ iterations of the \opl (or fewer if $N_T$ is hit earlier), however, at the end of each iteration do \emph{not} update the parent regardless of the outcome. Instead, for the $j$-th iteration we record the fitness $f'_j$ of the individual that would otherwise become the next parent.
\end{enumerate}

Note that the correctness of not updating the parent even in the case of neutral move is motivated by Requirement~\ref{algorithm-requirement}: any individual with the fitness $f$ induces the same behavior, so execution of a neutral move may be safely omitted.

Let $N_A \le N_I$ be the actual number of simulated iterations. The probability estimations $\tilde{p}_i$ of increasing the fitness by $i$ are computed as $\tilde{p}_i = \frac{1}{N_A} \cdot |\{ j \mid 1 \le j \le N_A; f'_j = f + i \}|$. Note that, as the \opl is elitist, $\tilde{p}_0$ counts also the occasions where the best offspring was worse than the parent.

Finally, $T_{f,p}$ is computed based on $\tilde{p}_i$ and $T^*_{f'}$ for $f' > f$ using the equation in line~\ref{dpTransitions} of Algorithm~\ref{alg:optimalAlg},
which is a solution of the recurrent relation $T_{f,p} = 1 + \tilde{p}_0 \cdot T_{f,p} + \sum_{i>0} \tilde{p}_i \cdot T^*_{f+i}$.

\subsection{Choice of Mutation Rate Sets}\label{sec:rate-sets}

The choice of the mutation rate sets $\{ p^{(f)}_1, \ldots, p^{(f)}_{m_f} \}$ has to be a trade-off between the accuracy of the resulting values and the computation time. However, the freedom to choose different sets for different fitness values partially reduces the effects of this trade-off. In particular, one may conduct preliminary experiments to see which probabilities for which fitness values are most promising, and increase the coverage near these probabilities in more involved experiments. Technically, this opens a possibility of a self-adaptive scheme, where first some predefined grid is used for each fitness value, and then it is refined in the most promising regions to get better results. In this paper, however, we used a more conservative scheme detailed below.

In our experiments, we used a union of two grids: the multiplicative grid of the form $p_i = p_{\text{base}} \cdot \alpha^i$, which we use for large fitness values and which spans relatively small probabilities,
and the additive grid of the form $p_i = p_{\text{base}} + i \cdot p_{\text{step}}$ to cover the whole ranges of fitness values and probabilities. As the particular parameter values need to depend on the problem size, we give these values, as well as the Monte Carlo simulation parameters $N_I$ and $N_T$, in the next section.

\LowerBoundPlot{!b}

\section{Example Application of Our Approach}
\label{sec:usecase}

In this section we outline several kinds of insights that can be derived from the results computed by the proposed algorithm.
Some of them, namely lower bounds for parameter control methods, plots of optimal parameter values, and parameter efficiency heatmaps,
have been previously proposed in~\cite{BuskulicD21,BuzdalovD20}, and the regret plots are new to this paper.


\subsection{The Ruggedness Function}
\label{sec:ruggedness}

We have chosen as use-case the \Ruggedness function introduced in~\cite{DoerrYHWSB20} (function F6 there). It is basically a function of concatenated jumps. That is, assuming that the unique optimum is located in some bit string $z$, with fitness value $n$, all points at Hamming distance one from $z$ have fitness $n-2$, while those at distance two have fitness $n-1$, those at distance three have fitness $n-4$, those at distance four have fitness $n-3$, and so on. Formally, letting $\OM_z(x):=\{ i \in [1..n] \mid z_i=x_i\}$, \Ruggedness assigns $r(x)=n$ if $\OM_z(x)=n$, $r(x)=\OM_z(x)+1$ if $\OM_z(x) \equiv n \mod 2$, and $r(x)=\OM_z(x)-1$ otherwise. 

As mentioned above, all the mutation rate control methods considered in~\cite{BuzdalovaDR20} perform worse on \Ruggedness than the shift \opl with static mutation rate $p=1/n$, see Fig.~\ref{fig:lower-bounds}. This raises two important questions, which we will answer in the remainder of this section:
(i) How far are the algorithms benchmarked in~\cite{BuzdalovaDR20} from the best possible performance? 
(ii) What is the impact of the individual parameter choices that are made during the optimization process? 

Our experiments consistently use problem size $n=100$. This choice is motivated by three main factors: it is large enough to see absolute differences in algorithms' behavior, yet small enough to finish the Monte Carlo simulations, and it allows a straightforward comparison with the previous works~\cite{BuzdalovaDR20,BuzdalovD20}.

To compute the necessary data for \Ruggedness, we use the following parameters for our Monte Carlo approach:
$N_I = 10^6$, $N_T = 5 \cdot 10^4$. We use the same set of probabilities for each fitness, which is constructed using a multiplicative
grid with the following parameters: $p_{\text{base}} = 10^{-4}$ to match $p_{\min}=1/n^2 = 10^{-4}$,
and $\alpha = 10^{1/25}$, so that 100 probabilities less than 1 are employed, whose logarithms are evenly distributed.

\subsection{Lower Runtime Bounds for Parameter Control Methods}

\begin{figure*}[!t]
    \centering
    \begin{tikzpicture}
    \begin{axis}[
    name=rates1,
	grid=both,
	width=0.51\textwidth,
	height=0.27\textheight,
	title={\OneMax},
	title style={at={(0.493,-0.25)}, anchor=north},
	xlabel={Fitness},
	ylabel={Scaled mutation rate: $p \cdot n$},
	ymode=log,
	legend columns = 5,
	legend style={at={(1.05,1.05)}, anchor=south},
	filter discard warning=false,
    enlarge y limits ={abs=1.5},
    cycle list name=colorcycle,
    ymin=0.1, ymax=100
]

    \addplot table [x=fitness, y=value, col sep=comma] {data/onemaxCSV/l2.csv};
	\addlegendentry{\footnotesize $\lambda = 2$}

    \addplot table [x=fitness, y=value, col sep=comma] {data/onemaxCSV/l4.csv};
	\addlegendentry{\footnotesize $\lambda = 4$}


	\addplot table [x=fitness, y=value, col sep=comma] {data/onemaxCSV/l16.csv};
	\addlegendentry{\footnotesize $\lambda = 16$}


	\addplot table [x=fitness, y=value, col sep=comma] {data/onemaxCSV/l64.csv};
	\addlegendentry{\footnotesize $\lambda = 64$}



	\addplot table [x=fitness, y=value, col sep=comma] {data/onemaxCSV/l512.csv};
	\addlegendentry{\footnotesize $\lambda = 512$}
\end{axis}
    \begin{axis}[
    name=rates2,
    grid=both,
    width=0.51\textwidth,
    height=0.27\textheight,
	title={\Ruggedness},
	at={(rates1.south east)}, xshift=1.2cm,
	title style={at={(0.493,-0.25)}, anchor=north},
	xlabel={Fitness},
	ymode=log,
	enlarge y limits ={abs=1.5},
	cycle list name=colorcycle,
	ymin=0.1, ymax=100
]
	
	\addplot table [x=fitness, y=mutationRate, col sep=comma] {data/allLambdasCsv/l2.csv};
	
	\addplot table [x=fitness, y=mutationRate, col sep=comma] {data/allLambdasCsv/l4.csv};
	
	
	\addplot table [x=fitness, y=mutationRate, col sep=comma] {data/allLambdasCsv/l16.csv};
	
	
	\addplot table [x=fitness, y=mutationRate, col sep=comma] {data/allLambdasCsv/l64.csv};
	
%
	
	\addplot table [x=fitness, y=mutationRate, col sep=comma] {data/allLambdasCsv/l512.csv};
\end{axis}
    \end{tikzpicture}
    \par
    \caption{Best mutation rates for different values of $\lambda$ and fitness values. Problem size is $n=100$ for both plots.}
    \label{fig:for-lambdas}
\end{figure*}
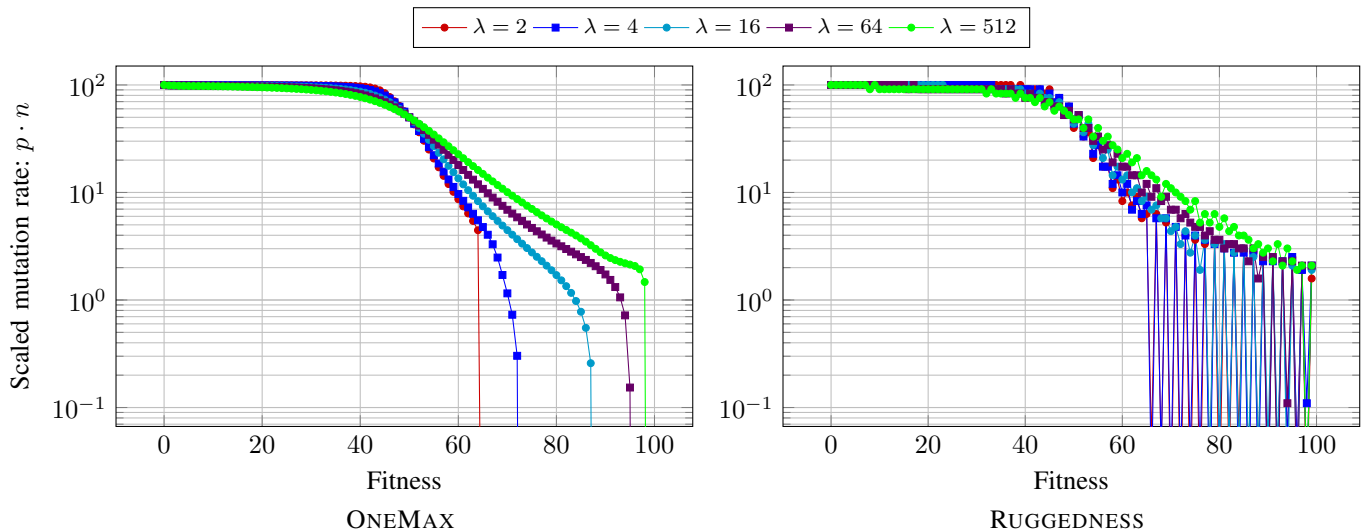    

The near-optimal runtime values $T^*_f$ for each fitness $f$ obtained from the proposed algorithm may be directly used to construct the lower bounds
on the expected running times possible for parameter control methods. For this, we compute for each fitness value $f$ the probability $p^{\text{init}}_f$ of hitting it with the first created individual.
The value $T = \sum_f T^*_f p^{\text{init}}_f$, which is the expected runtime of the \opl assuming near-optimal parameter choices, is then used as lower bound.

Fig.~\ref{fig:lower-bounds} presents the comparison of these lower bounds, measured in iterations, with the performance of the standard \opl, as well as of the \opl employing a few existing parameter control methods, on \Ruggedness and on \OneMax for comparison, all using the \shmut. For \OneMax we use the exact method for computing the lower bounds from~\cite{BuzdalovD20}. 
For the parameter control methods we additionally employ two different lower bounds on the mutation rate, $\pmn = 1/n$ and $\pmn = 1/n^2$.

Now we compare the insights achieved for \Ruggedness with the ones achieved before. In~\cite{BuzdalovaDR20} one was able to see only that some parameter control methods are far away from \opl when using $\pmn = 1/n^2$, and the situation becomes better when $\pmn = 1/n$ is used. More precisely, with the less generous lower bound on the mutation rate, all the considered methods are able to perform just as \opl. But is it possible to come up with a more efficient parameter control method that would outperform \opl? With the just obtained lower bounds on expected runtimes, represented with green lines in Fig.~\ref{fig:lower-bounds}, one can conclude that even if one makes the optimal mutation rate choices for each fitness value, only a relatively small constant-factor improvement would be possible. This gives an important bound which was not previously known. 

One of the interesting observations on \Ruggedness, which can be derived from Fig.~\ref{fig:lower-bounds}, is that the \AB has a much worse performance than both the \tworate and the \opl. However, the reason for such a behavior cannot be derived from Fig.~\ref{fig:lower-bounds}. We return to this question further in this section by looking more closely at how these methods choose mutation rates and how they relate to the best choices.

\subsection{Plots of Optimal Parameter Values}

The observation of running times alone cannot provide deeper insights about the problem structure, including the optimal mutation rates for different fitness values.
Such a knowledge can shed some light on which issues a parameter control method may face when the \opl solves the problem in question.
The plots of near-optimal mutation rates, which the proposed algorithm provides, may serve this purpose.
What is more, one can infer the influence of the population size $\lambda$ on the optimal parameters from such plots.

Fig.~\ref{fig:for-lambdas} presents the plots of near-optimal mutation rates as a function of the fitness value for \OneMax, computed as in~\cite{BuzdalovD20}, and \Ruggedness.
One can see that for \OneMax these plots demonstrate a steady tendency of decreasing while getting closer to the optimum,
which matches the general expectations. At the same time, most the plots for \Ruggedness, while having a similar general trend,
show regular oscillations with a period of 2. Indeed, a small mutation rate is better for the \opl to improve from
the fitness value that has a different parity than $n$ (by flipping a single bit), but at least two bits need to be flipped to improve from the fitness value with the same parity as $n$,
which requires a larger mutation rate that tends to $2/n$ as one gets closer to the optimum.

Note that such optimal parameters may be difficult to be tracked precisely by most parameter control methods that assume a slow change of the optimal mutation rate.


\begin{figure*}[!t]
    \BigHeatmapPlot
    \caption{Parameter efficiency heatmaps for \opl, $n=100, \lambda=512$.
             Heatmap for \Ruggedness also contains traces of mutation rate choices in runs of the \tworate (black) and the \AB (red), both with $p_{\min}=1/n^2$.
             Color denotes the relative efficiency, the larger the better.}
    \label{fig:heatmaps}
    \vspace{2ex}
	\begin{tikzpicture}
\begin{axis}[
    width=0.45\linewidth, height=0.254\textheight,
    grid=major,
    xlabel=Iteration number,
    ylabel=Regret in runtime,
    ymode=log,
    legend pos=south east,
    ymin=0.005,
    enlarge y limits = 0.1,
    enlarge x limits = 0,
    cycle list name=othercolorcycle
]
    \addplot coordinates {(0, 43.49915236949091)(1, 3.912918555183399)};
    \addplot coordinates {(1, 3.912918555183399)(2, 2.6582326007052)};
    \addplot coordinates {(2, 2.6582326007052)(3, 2.3769491081405723)};
    \addplot coordinates {(3, 2.3769491081405723)(4, 1.9170017483725363)};
    \addplot coordinates {(4, 1.9170017483725363)(5, 1.2635876110510404)};
    \addplot coordinates {(5, 1.2635876110510404)(6, 0.5143926723889546)};
    \addplot coordinates {(6, 0.5143926723889546)(7, 0.03573878317292135)};
    \addplot coordinates {(7, 0.03573878317292135)(8, 1.1879518500137773)};
    \addplot coordinates {(8, 1.1879518500137773)(9, 0.11696730470653692)};
    \addplot coordinates {(9, 0.11696730470653692)(10, 8.43738549649279E-4)};
    \addplot coordinates {(10, 8.43738549649279E-4)(11, 0.3159945520952713)};
    \addplot coordinates {(11, 0.3159945520952713)(12, 0.006792707023527016)};
    \addplot coordinates {(12, 0.006792707023527016)(13, 0.04519884909819821)};
    \addplot coordinates {(13, 0.04519884909819821)(14, 0.07078387486094183)};
    \addplot coordinates {(14, 0.07078387486094183)(15, 0.00602559946006632)};
    \addplot coordinates {(15, 0.00602559946006632)(16, 0.047657117100601276)};
    \addplot coordinates {(16, 0.047657117100601276)(17, 0.23098656951949964)(18, 0.23098656951949964)(19, 0.4318860435234381)};
    \addplot coordinates {(19, 0.4318860435234381)(20, 0.021109856747082562)};
    \addplot coordinates {(20, 0.021109856747082562)(21, 0.41183193063011775)(22, 0.08955311697554055)};
    \addplot coordinates {(22, 0.08955311697554055)(23, 1.2690705260195387)(24, 0.08955311697554055)(25, 3.144649023310971)};
    \addplot coordinates {(25, 3.144649023310971)(26, 0.06831957025720936)(27, 3.763218493218374)(28, 109.59219652799787)(29, 3.763218493218374)(30, 109.59219652799787)(31, 3.763218493218374)(32, 0.06831957025720936)(33, 3.763218493218374)(34, 0.1870757921986666)};
    \addplot coordinates {(34, 0.1870757921986666)(35, 17.553570616155714)(36, 17.553570616155714)(37, 0.1870757921986666)(38, 17.553570616155714)(39, 17.553570616155714)(40, 0.1870757921986666)(41, 29.508338830012715)(42, 0.1870757921986666)(43, 17.553570616155714)(44, 17.553570616155714)(45, 0.1870757921986666)(46, 17.553570616155714)(47, 0.1870757921986666)(48, 17.553570616155714)(49, 17.553570616155714)(50, 0.1870757921986666)(51, 29.508338830012715)(52, 0.1870757921986666)(53, 29.508338830012715)(54, 0.1870757921986666)(55, 17.553570616155714)(56, 0.1870757921986666)(57, 29.508338830012715)(58, 0.1870757921986666)(59, 17.553570616155714)(60, 17.553570616155714)(61, 0.1870757921986666)(62, 29.508338830012715)(63, 0.1870757921986666)(64, 29.508338830012715)(65, 0.1870757921986666)(66, 29.508338830012715)(67, 0.1870757921986666)(68, 29.508338830012715)(69, 0.1870757921986666)(70, 29.508338830012715)(71, 1346.3421525760969)(72, 29.508338830012715)(73, 0.1870757921986666)(74, 29.508338830012715)(75, 0.1870757921986666)(76, 29.508338830012715)(77, 0.1870757921986666)(78, 17.553570616155714)(79, 17.553570616155714)(80, 0.1870757921986666)(81, 29.508338830012715)(82, 0.1870757921986666)(83, 29.508338830012715)(84, 0.1870757921986666)(85, 29.508338830012715)(86, 0.1870757921986666)(87, 29.508338830012715)(88, 0.1870757921986666)(89, 29.508338830012715)(90, 1346.3421525760969)(91, 29.508338830012715)(92, 0.1870757921986666)(93, 17.553570616155714)(94, 0.1870757921986666)(95, 29.508338830012715)(96, 1346.3421525760969)(97, 29.508338830012715)(98, 0.1870757921986666)(99, 17.553570616155714)(100, 17.553570616155714)(101, 17.553570616155714)(102, 0.1870757921986666)(103, 29.508338830012715)(104, 0.1870757921986666)(105, 17.553570616155714)(106, 0.1870757921986666)(107, 29.508338830012715)(108, 0.1870757921986666)(109, 17.553570616155714)(110, 0.1870757921986666)(111, 29.508338830012715)(112, 0.1870757921986666)(113, 17.553570616155714)(114, 17.553570616155714)(115, 0.1870757921986666)(116, 29.508338830012715)(117, 0.1870757921986666)(118, 17.553570616155714)(119, 0.1870757921986666)(120, 29.508338830012715)(121, 1346.3421525760969)(122, 3029.269843296218)(123, 1346.3421525760969)(124, 29.508338830012715)(125, 0.1870757921986666)(126, 29.508338830012715)(127, 0.1870757921986666)(128, 29.508338830012715)(129, 0.1870757921986666)(130, 29.508338830012715)(131, 0.1870757921986666)(132, 29.508338830012715)(133, 0.1870757921986666)(134, 29.508338830012715)(135, 0.1870757921986666)(136, 17.553570616155714)(137, 0.1870757921986666)(138, 29.508338830012715)(139, 1346.3421525760969)(140, 29.508338830012715)(141, 0.1870757921986666)(142, 29.508338830012715)(143, 0.1870757921986666)(144, 17.553570616155714)};
    \addlegendentry{\footnotesize \rate, $\lambda = 512$};
    \draw[red,ultra thick, domain=0:146] plot (\x, 2019.5132288641453);
\end{axis}
\end{tikzpicture}\hspace{1em}
	\begin{tikzpicture}
\begin{groupplot}[
    group style={
        group size=5 by 1,
        yticklabels at=edge left,
        horizontal sep=3pt
    },
    height=0.254\textheight,
    grid=major,
    ymode=log,
    ymin=0.005,
    enlarge y limits = 0.1,
    ymin=0.01, ymax=130000,
    cycle list name=othercolorcycle,
    legend style={at={(0.5,0.025)},anchor=south east}
]
    \nextgroupplot[xmin=0, xmax=60, width=0.3\linewidth, ylabel=Regret in runtime, xlabel=Iteration number, xtick={0,20,40,60}]
    \addplot coordinates {(0, 43.49915236949091)(1, 3.912918555183399)};
    \addplot coordinates {(1, 3.912918555183399)(2, 3.2178308666475273)};
    \addplot coordinates {(2, 3.2178308666475273)(3, 2.835877990195634)};
    \addplot coordinates {(3, 2.835877990195634)(4, 1.821287805681154)};
    \addplot coordinates {(4, 1.821287805681154)(5, 0.4793412292461008)};
    \addplot coordinates {(5, 0.4793412292461008)(6, 0.13152814797568174)};
    \addplot coordinates {(6, 0.13152814797568174)(7, 2.473101481001916)};
    \addplot coordinates {(7, 2.473101481001916)(8, 11.574291672248544)};
    \addplot coordinates {(8, 11.574291672248544)(9, 59.809874196133464)};
    \addplot coordinates {(9, 59.809874196133464)(10, 59.809874196133464)(11, 1.1161988757779642)};
    \addplot coordinates {(11, 1.1161988757779642)(12, 4467.151422132555)(13, 2.831522210309262)};
    \addplot coordinates {(13, 2.831522210309262)(14, 16045.5539510481)(15, 7.359363391687132)};
    \addplot coordinates {(15, 7.359363391687132)(16, 112487.0788085341)(17, 7.359363391687132)(18, 0.3159945520952713)};
    \addplot coordinates {(18, 0.3159945520952713)(19, 21.997454404453453)(20, 0.6392298810129666)};
    \addplot coordinates {(20, 0.6392298810129666)(21, 66.88732691333057)(22, 1.5362571706416466)};
    \addplot coordinates {(22, 1.5362571706416466)(23, 244.0701165426945)(24, 3.466507730512093)};
    \addplot coordinates {(24, 3.466507730512093)(25, 945.6657577216496)(26, 19.62059864211715)};
    \addplot coordinates {(26, 19.62059864211715)(27, 47516.38173055398)(28, 19.62059864211715)(29, 0.611351430320477)(30, 0.018697882140788463)};
    \addplot coordinates {(30, 0.018697882140788463)(31, 1.3619480801480999)(32, 0.018697882140788463)(33, 0.08640502903661082)(34, 0.7696958455832856)(35, 3.2663537437974477)(36, 20.9598410476846)};
    \addplot coordinates {(36, 20.9598410476846)(37, 5.447538823651577)(38, 20.9598410476846)(39, 78.58285955871916)(40, 273.09232529944916)(41, 1143.1610334066431)(42, 3921.0119251510578)(43, 4998.370130869366)(44, 4998.370130869366)(45, 4998.370130869366)(46, 4998.370130869366)(47, 4998.370130869366)(48, 4998.370130869366)(49, 4998.370130869366)(50, 4998.370130869366)(51, 4998.370130869366)(52, 4998.370130869366)(53, 4998.370130869366)(54, 4998.370130869366)(55, 4998.370130869366)(56, 4998.370130869366)(57, 4998.370130869366)(58, 4998.370130869366)(59, 4998.370130869366)(60, 4998.370130869366)};
    \addplot plot[red, ultra thick, mark size=0pt] coordinates {(0,74991.38587235607)(60,74991.38587235607)};

    \nextgroupplot[xmin=9700, xmax=9715, width=0.135\linewidth, ticks=none, cycle list name=othercolorcycleinv]
    \addplot coordinates {(9700, 4998.370130869366)(9701, 4998.370130869366)(9702, 4998.370130869366)(9703, 4998.370130869366)(9704, 4998.370130869366)(9705, 4998.370130869366)(9706, 4998.370130869366)(9707, 12497.431251424638)};
    \addplot coordinates {(9707, 12497.431251424638)(9708, 4592.648292531737)(9709, 12497.431251424638)(9710, 12497.431251424638)(9711, 12497.431251424638)(9712, 12497.431251424638)(9713, 12497.431251424638)(9714, 12497.431251424638)(9715, 12497.431251424638)};
    \addplot plot[red, ultra thick, mark size=0pt] coordinates {(9700,74991.38587235607)(9715,74991.38587235607)};

    \nextgroupplot[xmin=32030, xmax=32042, width=0.13\linewidth, ticks=none]
    \addplot coordinates {(32030, 12497.431251424638)(32031, 12497.431251424638)(32032, 12497.431251424638)(32033, 12497.431251424638)(32034, 12497.431251424638)(32035, 12497.431251424638)(32036, 49994.25724823738)};
    \addplot coordinates {(32036, 49994.25724823738)(32037, 7137.11439114308)(32038, 49994.25724823738)(32039, 49994.25724823738)(32040, 49994.25724823738)(32041, 49994.25724823738)(32042, 49994.25724823738)};
    \addplot plot[red, ultra thick, mark size=0pt] coordinates {(32030,74991.38587235607)(32042,74991.38587235607)};

    \nextgroupplot[xmin=43845, xmax=43860, width=0.14\linewidth, ticks=none, cycle list name=othercolorcycleinv]
    \addplot coordinates {(43845, 49994.25724823738)(43846, 49994.25724823738)(43847, 49994.25724823738)(43848, 49994.25724823738)(43849, 49994.25724823738)(43850, 49994.25724823738)(43851, 49994.25724823738)(43852, 49964.36208120445)};
    \addplot coordinates {(43852, 49964.36208120445)(43853, 37820.87331457144)(43854, 49964.36208120445)(43855, 49964.36208120445)(43856, 49964.36208120445)(43857, 49964.36208120445)(43858, 49964.36208120445)(43859, 49964.36208120445)(43860, 49964.36208120445)};
    \addplot plot[red, ultra thick, mark size=0pt] coordinates {(43845,74991.38587235607)(43860,74991.38587235607)};

    \nextgroupplot[xmin=53140, xmax=53144, width=0.1\linewidth, xtick={53144}, scaled x ticks=false]
    \addplot coordinates {(53140, 49964.36208120445)(53141, 49964.36208120445)(53142, 49964.36208120445)(53143, 49964.36208120445)(53144, 49964.36208120445)};
    \addplot plot[red, ultra thick, mark size=0pt] coordinates {(53140,74991.38587235607)(53144,74991.38587235607)};

    \addlegendentry{\footnotesize \AB, $\lambda = 512$}
\end{groupplot}
\end{tikzpicture}
    \vspace{-4ex}
    \caption{Difference between the expected runtimes for the best mutation rate choice and the one actually chosen by the parameter control algorithms in example runs.
             Left plot shows a complete run of the \tworate. Right plot shows selected iterations of the \AB, the omitted iterations have the same fitness and regret.
             Each red line denotes an artificial visual boundary between finite expected runtimes (below) and infinite ones (above).}
    \label{fig:regrets}
\end{figure*}

\subsection{Parameter Efficiency Heatmaps}

Since Algorithm~\ref{alg:optimalAlg} provides expected runtimes $T_{f,p}$ for each fitness value and mutation rate, assuming that for higher fitness values the optimal mutation rates are chosen, we can use this data to estimate the relative efficiency of mutation rate choices for each fitness value. In Fig.~\ref{fig:heatmaps} we present this information as a heatmap for $n=100$ and $\lambda=512$. Each cell of a heatmap corresponds to a pair of $f$ and $p$, where the following $p$ are chosen: $p = 10^{-4 + i/25}$ for $0 \le i \le 100$.

Colors of the cells represent the relative efficiency of the corresponding $p$ among all mutation rates for the corresponding $f$. They are derived from the value $C_{f,p} = \exp(\alpha_f \cdot (T^*_f - T_{f,p}))$, where $\alpha_f \le 1$ is chosen in such a way that at least a half of $C_{f,p}$ are at least $0.5$. This is done to display enough information about at least 50\% of possible choices, while not artificially emphasizing differences if they are negligible.

Using Fig.~\ref{fig:heatmaps}, one can make the following observations about dynamic mutation rates for the \opl on \Ruggedness:
\begin{itemize}
\item Overall, near-optimal mutation rates start with a rather high value at the beginning of the optimization and show a trend to gradually decrease towards the optimum.
\item This trend is non-monotone. By comparison with \OneMax, one can see that for even fitness values \Ruggedness shows behavior similar to \OneMax.
      However, for odd fitness values the best mutation rates are higher.
\end{itemize}

These observations agree well with the findings from the previous subsection and Fig.~\ref{fig:for-lambdas}. However, one can also see that in most of fitness values, except for a few last ones,
there is a wide enough range of nearly equally good parameters for each fitness value, which shows that most parameter control methods have, in theory, a fair chance of maintaining good mutation rates
until the last moments.

Fig.~\ref{fig:heatmaps} also shows traces of the actual mutation rates chosen by the \tworate and \AB algorithms on \Ruggedness, both with $p_{\min}=1/n^2$,
for the corresponding fitness values. The following observations can be made:
\begin{itemize}
\item Both algorithms quickly, in less than 10 iterations, reach near-optimal mutation rates at fitness of roughly 70.
\item For fitness of roughly 80, both algorithms choose reasonably good values often enough, which ensures some progress, with the \tworate doing slightly better.
\item For last few fitness values, both algorithms visit odd fitness values exclusively, but the \tworate still chooses good mutation rates, whereas the \AB gets stuck with the too small rates. 
\end{itemize}

The latter observation clearly points at the behavior that causes the poor performance of the \AB seen at Fig.~\ref{fig:lower-bounds}.
It appears that the $(A,b)$ adaptation rule forces a switch to local search as progress slows down,
which converges the mutation rate to $p_{\min}=1/n^2$. With such rate it makes the chance of finding two-bit improvements $\Theta(n^2)$ smaller,
which explains the large performance gap observed in Fig.~\ref{fig:lower-bounds}.


\subsection{Regret Plots}

Overlaying parameter choices of particular parameter control methods over the parameter efficiency heatmaps may indicate the presence of problems in a method, and how large the deviation is. However, it cannot show how much of the performance the method loses from acting suboptimally.

A possible solution is to show how much the method loses to the optimal choice in each iteration. More precisely, for each iteration we take the parameter value $p$ chosen by the parameter control method in question that corresponds to the current parent's fitness value, and plot the value $|T_{f,p} - T^*_{f}|$ for that iteration. Such \emph{regret plots} are presented in Fig.~\ref{fig:regrets}. We change the plot color whenever the fitness value changes, so that one can see which iterations share the same fitness value.

We can observe in the example runs in Fig.~\ref{fig:regrets} that both methods initially quickly reduce their regret. However, the \tworate keeps it at a low level, whereas the \AB shows a more complicated behavior. First it features two peaks of local regret maxima, which match two occasions of too high mutation rates seen in Fig.~\ref{fig:heatmaps}. It then quickly moves through a good region and falls down to small mutation rates, where it spends most of its time with very large regrets, whose sum dominates the total running time.

Note that, occasionally, a parameter control method may enter a region where $T_{f,p}$ reaches too high values, or where it is even infinite (which happens twice in Fig.~\ref{fig:regrets}). If the method is quick enough to return to better mutation rates, this does not harm the performance, as is the case for the \tworate.


%

\section{Generalization of the Hybrid Approach}
\label{sec:generalization}

As discussed in the introduction, our hybrid approach is at the moment restricted to settings in which states are not visited more than once. This can be resolved, at the price of computational complexity, by construction of Markov chains on all states with equal fitness and solving the resulting system of equations. However, this may also require a revision of used definitions, such as the one for regret: if the mutation rate is fixed to some value for one of the states, globally \emph{suboptimal} choices for other states may actually make the runtime \emph{smaller}.

Our approach is, in fact, not limited to $(1+\lambda)$ type algorithms: as it simulates an iteration of the algorithm as a parameterized black-box, a wide range of algorithms may be modeled, which is much harder with exact approaches.

The number of states may become too large to have them considered explicitly for certain problems, as well as for population-based algorithms. While we see no obvious solution for this problem yet,
state-merging approaches might be a solution for some of the cases.

\section{Conclusions}
\label{sec:conclusions}

We have extended in this work the dynamic programming approach for computing optimal state-dependent, dynamic mutation rates suggested in~\cite{BuskulicD21,BuzdalovD20} to settings in which the transition probabilities cannot necessarily be computed by  closed-form expressions, but where they need to be approximated by Monte Carlo simulations. We have applied this approach to derive optimal parameter choices for the \opl on \Ruggedness. We have also introduced \emph{regret plots}, which not only show deficiencies in parameter control methods, but indicate their impact on the running time.

We plan to extend our work to more complicated setting detailed in Section~\ref{sec:generalization}, as well as to use the results presented in this work to improve some of the parameter control methods.

\vspace{1ex}
{{\textbf{Acknowledgments.} 
The reported study was funded by RFBR and CNRS, project number 20-51-15009, and by the Paris Ile-de-France region.
}}


\end{document}